\documentclass[journal, twoside]{IEEEtran}

\usepackage{cite}
\usepackage{amsmath}
\usepackage{amsfonts}
\usepackage{array}
\usepackage{booktabs}
\usepackage{algorithm}
\usepackage[noend]{algpseudocode}
\usepackage{color}
\usepackage[caption=false,font=footnotesize]{subfig}
\usepackage{url}
\usepackage[pdftex]{graphicx}
\usepackage{hyperref}
\usepackage{bbm}
\usepackage[normalem]{ulem}
\usepackage{multirow}

\newcommand{\RR}{\ensuremath{\mathbb R}}
\newcommand{\minimize}[2]{\ensuremath{\underset{\substack{{#1}}}{\operatorname{minimize}}\;\;#2}}
\newcommand{\argmin}[2]{\ensuremath{\underset{\substack{{#1}}}{\operatorname{argmin}}\;\;#2}}
\newcommand{\diag}{\ensuremath{\operatorname{diag}}}
\newcommand{\Tr}{\ensuremath{\operatorname{Tr}}}
\newcommand{\Log}{\operatorname{Log}}
\newcommand{\Exp}{\operatorname{Exp}}
\newcommand{\chol}{\operatorname{Cholesky}}

\begin{document}

	\title{OrthoNet: Multilayer Network Data Clustering}
	
	\author{Mireille~El~Gheche,
		Giovanni~Chierchia,
		and~Pascal~Frossard% <-this % stops a space
		\thanks{Mireille El Gheche and Pascal Frossard are with Signal Processing Laboratory (LTS4), Ecole Polytechnique F\'ed\'erale de Lausanne (EPFL), Lausanne, Switzerland. E-mail: mireille.elgheche@epfl.ch, pascal.frossard@epfl.ch}% <-this % stops a space
		\thanks{Giovanni Chierchia is with Universit\'e Paris-Est, LIGM UMR 8049, CNRS, ESIEE Paris, Noisy-le-Grand, France. E-mail: giovanni.chierchia@esiee.fr}% <-this % stops a space
		\thanks{Giovanni Chierchia was supported by the French CNRS INS2I with a PEPS JCJC project funded under the grant 2019OSCI.}%
	}

	\markboth{IEEE TRANSACTIONS ON SIGNAL AND INFORMATION PROCESSING OVER NETWORKS}%,~Vol.~14, No.~8, August~2015}%
	{El Gheche \MakeLowercase{\textit{et al.}}: OrthoNet: Multilayer Network Data Clustering}
	
	

	\maketitle
	
	%--------------------------------------------------------------
	\begin{abstract}
		%--------------------------------------------------------------
		%
		%
		%
		Network data appears in very diverse applications, like biological, social, or sensor networks. Clustering of network nodes into categories or communities has thus become a very common task in machine learning and data mining. Network data comes with some information about the network edges. In some cases, this network information can even be given with multiple views or layers, each one representing a different type of relationship between the network nodes. Increasingly often, network nodes also carry a feature vector. We propose in this paper to extend the node clustering problem, that commonly considers only the network information, to a problem where both the network information and the node features are considered together for learning a clustering-friendly representation of the feature space. Specifically, we design a generic two-step algorithm for multilayer network data clustering. The first step aggregates the different layers of network information into a graph representation given by the geometric mean of the network Laplacian matrices. The second step uses a neural net to learn a feature embedding that is consistent with the structure given by the network layers. We propose a novel algorithm for efficiently training the neural net via gradient descent, which encourages the neural net outputs to span the leading eigenvectors of the aggregated Laplacian matrix, in order to capture the pairwise interactions on the network, and provide a clustering-friendly representation of the feature space.
		We demonstrate with an extensive set of experiments on synthetic and real datasets that our method leads to a significant improvement w.r.t.\ state-of-the-art multilayer graph clustering algorithms, as it judiciously combines nodes features and network information in the node embedding algorithms.
	\end{abstract}
	

	%--------------------------------------------------------------
	\section{Introduction}
	%--------------------------------------------------------------
	%
	%
	%
	\IEEEPARstart{N}{etwork} data is getting increasingly popular in machine learning and data science, as it corresponds to a natural data representation form in biological, social, computer, or sensor network applications, to cite a few examples. Network data can be mathematically described by graphs whose vertices and edges correspond to the network nodes and links respectively. Even in applications where the network information is not explicit, graphs can be used to model the pairwise relationships between data points. Moreover, applications often rely on multiple sources of information to characterize the relationships between data. This leads to multilayer network representations, where nodes are shared across network layers, each one describing a different type of relationship between network nodes. In addition, it is often possible to associate attributes with the network nodes, which may represent different forms of measurements or signals. For example, a public transport system can be represented by a multilayer graph, where the nodes are transportation hubs, each layer describes a different mean of transportation (a bus line, a metro line, etc.), and the node attribute is the number of travelers at each hub. This obviously leads to very rich datasets, and it becomes important to devise machine learning methods that are able to consider altogether the information of both the multilayer network and the node features. 

Multilayer networks are considered in many machine learning and data mining tasks, including inference of mixture models, multi-view learning, processing, clustering, and community detection. We focus here on the multilayer network data clustering problem, where the goal is to assign each network node (shared across different layers) to a cluster, by taking into account both the feature vectors on the nodes and the connectivity patterns in each layer. Multilayer network data clustering differs from common classes of clustering methods in two main aspects: (i) the information about cluster membership must be estimated from multiple network layers, while classical network clustering only considers a single layer; (ii) the clusters are formed by considering both node features and network information, while graph clustering algorithms usually only rely on network information.

In this paper, we present a general approach for multilayer network data clustering, which exploits both the Riemannian geometry of the symmetric positive definite (SPD) manifold and the power of neural nets to learn a proper node embedding. Given the intrinsic difficulty of jointly considering both node features and network information, we propose a constructive solution that works in two consecutive steps. Firstly, we compute the geometric mean of Laplacian matrices associated to each layer of the network. Aggregating the multilayer network into a graph representation with the form of a SPD matrix allows us to properly take into account the topology shared across layers. Secondly, we use the aggregated SPD matrix and the node features to perform deep spectral clustering. Unlike the standard approach of Laplacian matrix eigendecomposition, we reformulate spectral clustering as a trace optimization problem subject to an orthogonality constraint, and we devise a new algorithm to solve it. The peculiarity of our approach is that the orthogonality constraint is enforced implicitly, leading to a differentiable cost function that can be optimized via gradient descent. This allows us to use a neural net for learning the node feature embedding, which is thereby trained without supervision. Similarly to spectral clustering, the goal is to find a nonlinear mapping of the node features that penalizes the pairwise interactions provided by the aggregated SPD matrix, while enforcing the orthogonality constraint in the low-dimensional space to avoid trivial solutions.

Experimental results on diverse datasets show that the proposed approach has a better clustering performance compared to baseline multilayer network data clustering approaches, due to the effective combination of network and node feature information. We expect that our algorithm can provide a new generic solution for the effective processing of rich network datasets with combinations of different forms of information.

The remaining of this paper is organized as follows. Section \ref{sec:relatedwork} reviews the literature on multilayer network data clustering. Section \ref{sec:problem} describes the problem formulation. Section \ref{sec:embedding} details our approach for node feature embedding based on a learning objective inspired from spectral clustering. Section \ref{sec:spd} presents our approach for layer aggregation based on the geometric mean of SPD matrices. Section \ref{sec:results} provides an experimental validation of the proposed approach on synthetic and real multilayer graphs. Section \ref{sec:conclusion} draws the conclusion.

%-----------------------
\section{Related work}
\label{sec:relatedwork}
%-----------------------
A wide panel of approaches were proposed to combine the information from multilayer networks, and an intense research effort was dedicated to clustering methods. In this section, we review the literature on multilayer network aggregation and graph representation learning.

\subsection{Multilayer network aggregation}
The most straightforward way to summarize the information from a multilayer graph is to perform a linear or convex combination of its layers \cite{NIPS2005_2938, Tang_20012, Chen_Hero_2017}. While convex combinations can be efficient in some cases, they may not be able to capture the specificity present in each layer. In this regard, more effective ways to merge the graph layers is to make use of the family of matrix power mean \cite{MercadoGT018}, or to see them as points of a Grassmann manifold \cite{Wang_TIP_2013}. 

A different aggregation strategy consists of integrating the information from individual layers directly into the optimization problem underlying the learning process. Examples include the co-EM clustering algorithm \cite{Bickel2004}, the clustering approach in \cite{Kumar2011} based on co-training \cite{Blum1998} and co-regularization \cite{Sindhwani2005}, as well as the joint fusion and clustering approach in \cite{Wang2019}. These methods can be useful when a unified representation for the multiple views is not easy to find in the data. In \cite{Dong_TSP_2014}, each graph layer is modeled as a subspace on a Grassman manifold, and they are combined by finding the subspace that minimizes the sum of projection distances to all layers.  

Closer to this paper, the work in \cite{Dong_TSP_2014} performs the aggregation in the Grassmann manifold. However, it lacks a meaningful summarization of the information contained in graph layers, and neglects any attribute that may be assigned to the graph nodes. The main novelty of the proposed approach w.r.t.\ \cite{Dong_TSP_2014} lies in the introduction of a new numerical algorithm to combine the characteristics of graph layers in the SPD manifold, and the design of a new approach that takes into account features carrying relevant information about the nodes.

\subsection{Graph representation learning}
In the study of graphs and networks, community detection refers to the problem of grouping together nodes that are more densely connected internally than with the rest of the network \cite{clauset2008, Fortunato2010, kim_2015_acm_sigmod, leo_2015_statistical}. In this paper, we are mainly interested in graph clustering based on spectral analysis \cite{Schaeffer2007}. Spectral clustering can be linked to dimensionality reduction, which aims at representing graphs and/or high-dimensional data into low-dimensional spaces (also called embedding), while preserving both the graph topological structure and the node content information. In this regard, one of the most popular techniques consists of embedding the graph nodes into the subspace spanned by the eigenvectors of the graph Laplacian matrix corresponding to the $K$ smallest eigenvalues \cite{Malik2000,Ng2001}, where clusters can be easily detected via K-means algorithm \cite{Macqueen1967}. Extensions of this approach consider the introduction of suitable constraints into the problem formulation, with the aim of conveying some prior knowledge on the cluster analysis \cite{Xu2000,Wang2010}. Alternatively, one can use the first $K$ eigenvectors of the graph Laplacian matrix in a modularity maximization problem \cite{Newman2006,2Newman2006}. Another approach hinges around the interpretation of Principal Component Analysis (PCA) on graphs \cite{Saerens2004}, which again links the graph structure to a subspace spanned by the top eigenvectors of the graph Laplacian matrix. Moreover, numerous methods have been proposed in the literature for representation learning on graphs, such as multidimentional scaling (MDS) \cite{Kruskal1978}, Laplacian eigenmap \cite{Belkin2001}, IsoMap \cite{Tenenbaum2000}, LLE \cite{Roweis2000}, matrix factorization \cite{Yang2018,Shen2018}, random walks \cite{Perozzi2014}, and deep learning approaches \cite{hamilton2017, Shahaman2018, bojchevski2018deep, Pfau2019}.

The works in \cite{Shahaman2018, Pfau2019} propose to learn a nonlinear map that embeds data points into the eigenspace of their associated graph Laplacian matrix, and subsequently clusters them. Differently from \cite{Shahaman2018, Pfau2019}, we use a multilayer graph signal, and we propose a new algorithm for learning the nonlinear map. In this respect, the originality of our approach lies in the reformulation of the optimization problem, in which we replace the orthogonality constraint with a differentiable operation injected directly into the cost function. In this regard, the main advantage of our approach is to avoid the complexity of alternating between a projection step and a gradient step like in \cite{Shahaman2018}, as the latter may slow down the training process.

	%--------------------------------------------------------------
	\section{OrthoNet Framework}
	%\section{Problem Formulation}
	\label{sec:problem}
	%--------------------------------------------------------------
	%
	%
	%
	
\subsection{Problem Formulation}

\begin{figure*}
	\centering
	\subfloat[Features
	\label{fig:toy:features}]{\includegraphics[width=0.18\linewidth, trim= 0 0 0 4.8cm, clip]{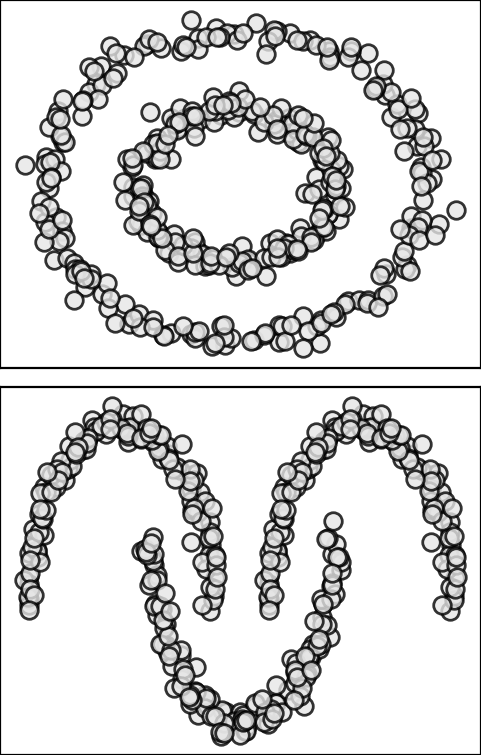}}
	\hfill
	\subfloat[Nearest-neighbor graph \label{fig:toy:graph}]{\includegraphics[width=0.18\linewidth, trim= 0 0 0 4.8cm, clip]{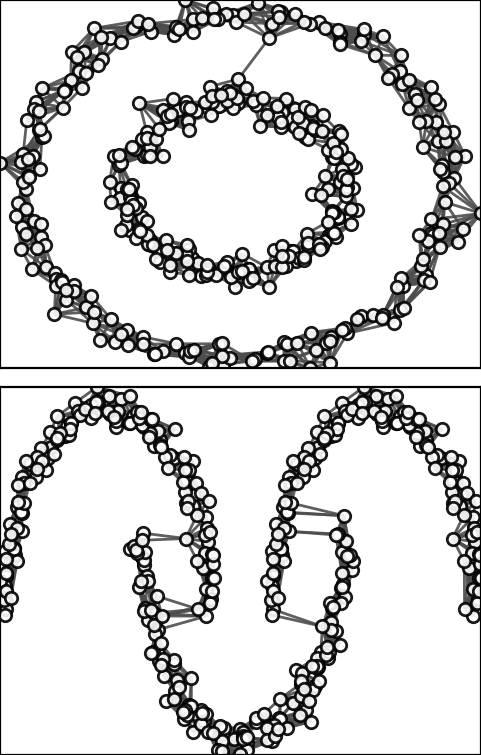}}
	\hfill
	\subfloat[Feature embedding \label{fig:toy:embedding}]{\includegraphics[width=0.18\linewidth, trim= 0 0 0 4.8cm, clip]{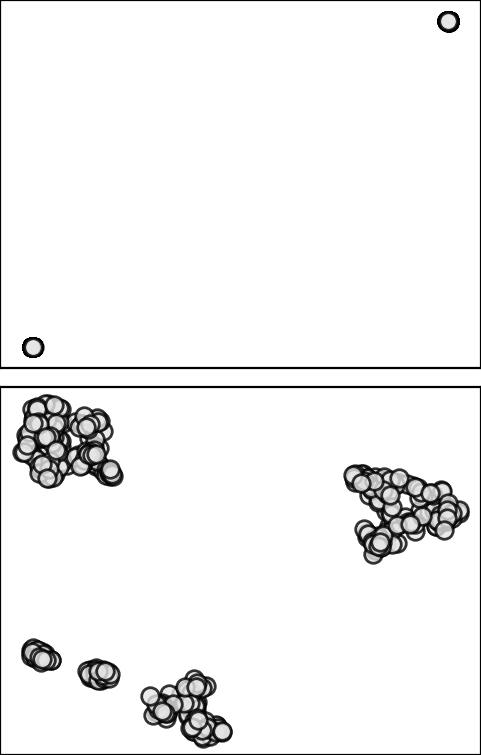}}
	\hfill
	\subfloat[K-means clustering (stars mark the cluster centers) \label{fig:toy:clustering}]{\includegraphics[width=0.18\linewidth, trim= 0 0 0 4.8cm, clip]{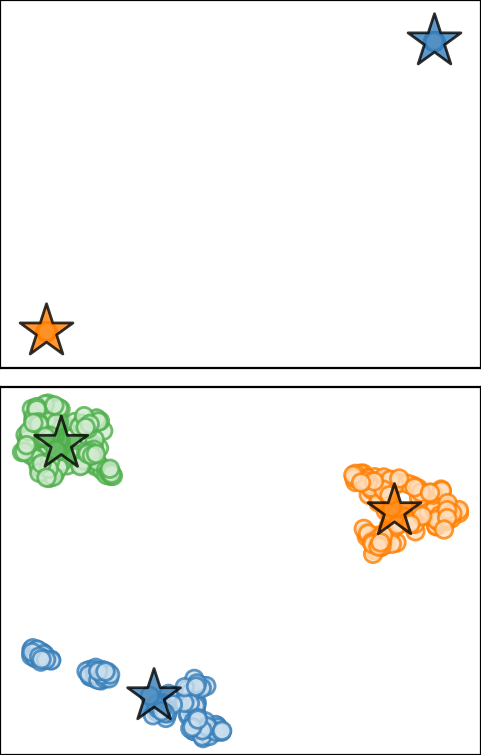}}
	\hfill
	\subfloat[Classification (lines mark the decision boundaries) \label{fig:toy:classification}]{\includegraphics[width=0.18\linewidth, trim= 0 0 0 4.8cm, clip]{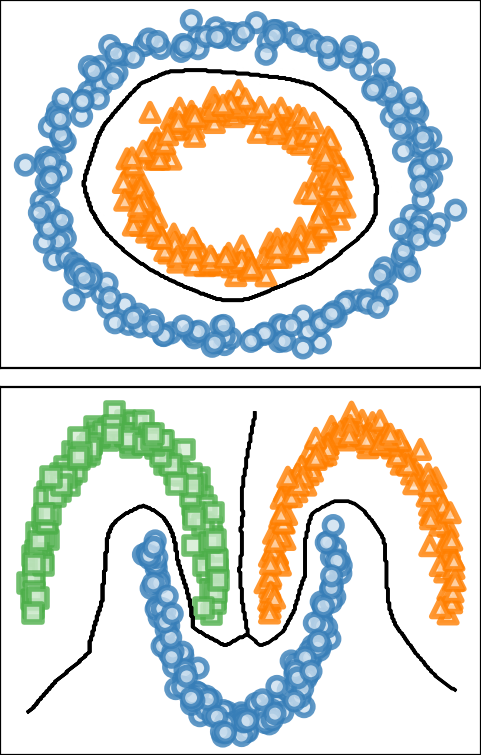}}
	\caption{Illustration of the proposed framework. Given the features (a) and the graph (b) representing their interactions, a mapping is learned so as to transform the features of strongly connected nodes into close vectors of a latent space (c). Doing so, the mapped features yield a representation that can be easily clustered with K-means (d). In addition, as the learned mapping can be applied to any feature vector, it is possible to define a classifier (e) that takes its decision based on the distance of a latent vector to the cluster centers computed on the feature embedding. The illustration is given for  $N=500, M=2, K=3, S=1$.}
	\label{fig:toy}
\end{figure*}

We are interested in clustering a multilayer graph
\begin{equation}
\mathcal{G} = \big\{\mathcal{G}^s(V,E^s)\big\}_{1\le s\le S}
\end{equation}
defined on a set $V$ of $N$ vertexes shared across $S \geq 1$ layers of edges.\footnote{In this paper, the terms graph and network, vertex and node, as well as multilayer, multiview, and multiplex are used interchangeably.} 
For each layer $s\in\{1,\dots,S\}$, there is a graph $\mathcal{G}^s(V,E^s)$ with (non-negative) similarity edge weights. We denote by $W^s = [w^s_{i,j}] \in \mathbb{R}^{N\times N}$ the weighted adjacency matrix of $\mathcal{G}^s$, and by $D^s=\diag(d_1^s,\dots,d_N^s)$ the diagonal matrix of vertex degrees $d_i^s = \sum_j w^s_{ij}$ for all $i\in\{1,\dots,N\}$. The Laplacian matrix of $\mathcal{G}^s$ is thus defined as
\begin{equation}
L^s=D^s-W^s. 
\end{equation}
We further assume that each vertex of the multilayer graph $\mathcal{G}$ is associated with $M$-dimensional features, and we denote such network data as
\begin{equation}
X = 
\begin{bmatrix}
x_1^\top\\ 
\vdots\\ 
x_N^\top
\end{bmatrix} 
\in \mathbb{R}^{N\times M}.
\end{equation}

Our goal is to cluster the graph vertices by taking into account both the multilayer structure of $\mathcal{G}$ and the node features $X$, without any a priori information about the actual relationship between the graph layers and the features. We however rely on three minimal assumptions, listed below.
\begin{enumerate}
    \item \textbf{Node connectivity.} Nodes that are connected in multiple graph layers are more likely to belong to the same cluster.
    \item \textbf{Layer complementarity.} Individual layers provide at least a partial information on the clustering structure.
    \item \textbf{Feature correlation.} Features for nodes within the same clusters are likely to be more correlated than features for nodes in different clusters. 
    \end{enumerate}
This setting is especially useful in scenarios where the topology shared across layers provides information that is not fully present neither in the data, nor in each layer alone.

As we do not assume any a priori model between the graph layers and the node features, we aim at discovering their relationship from the data. To this end, we propose a learning approach that exploits the topology shared across layers to drive the unsupervised training of a feature mapping.  We define a learning objective that encourages the features of strongly connected nodes to be mapped to close vectors of a latent space. By doing so, the learned mapping bears similarity with spectral clustering, and yields a clustering-friendly representation of the node features. 

In cases where node connectivity implies feature correlation (see our assumptions), the mapping actually learns to represent correlated features as close vectors of the latent space, thus yielding a clustering-friendly representation of the whole feature space. This makes it possible to obtain a classifier that generalizes to any feature vector, included but not limited to those associated with the graph nodes.

The proposed learning framework can be formulated as the joint optimization problem of finding the graph that is representative of all layers, and the mapping that allows for the clustering of its nodes. In general, this task is complex to solve, especially since we have no assumption on the interactions between graph layers and node features. We present a constructive solution to this problem in the next section.  

\subsection{Proposed approach}
Our approach is based on the idea of using the multilayer graph information, especially the information that appears consistently across layers, to drive the unsupervised learning of a mapping on the node features. Given the intrinsic difficulty of this joint optimization problem, we propose an alternative solution in two consecutive steps, detailed in the following.

\begin{enumerate}
	\item \textbf{Layer aggregation}. In the first step, we merge the individual layers into a representative graph $G$ given by its Laplacian matrix $L$. This operation is performed directly on the Laplacian matrices through an operator $\Phi\colon(\RR^{N\times N})^S\to\RR^{N\times N}$, that is
	\begin{equation}
	L = \Phi(L^1, \dots, L^S).
	\end{equation}
	 We propose to compute $L$ as the geometric mean (in the SPD manifold) of the Laplacian matrices $L^s$ given by the layers $\mathcal{G}^s$. This allows us to summarize the topology shared across layers into a single Laplacian matrix, as we shall explain in Section \ref{sec:spd}.
	
	\item \textbf{Feature embedding}. In the second step, we estimate the parameters $\theta\in\RR^B$ of a nonlinear mapping defined as
	\begin{equation}
	f_\theta\colon\mathbb{R}^{M}\to\mathbb{R}^{K},
	\end{equation} 
 	which embeds the node features in a latent space of dimension equal to the number $K$ of desired clusters. We perform this task using a learning objective inspired from spectral clustering, where the representative graph given by the Laplacian matrix $L$ drives the unsupervised learning of the mapping on the node features, as we will present in Section~\ref{sec:embedding}.
	
\end{enumerate}

Once the mapping $f_\theta$ has been learned from the multilayer graph, the node features are transformed as
\begin{equation}\label{eq:Y_theta}
Y_\theta = f_\theta(X) = 
\begin{bmatrix}
f_\theta(x_1)^\top\\ 
\vdots\\ 
f_\theta(x_N)^\top
\end{bmatrix}
\in \mathbb{R}^{N\times K}.
\end{equation}
The matrix $Y_\theta$ provides a clustering-friendly representation of the graph nodes. This is ensured by our learning objective, which encourages the mapping $f_\theta$ to represent the features of strongly connected nodes as close vectors of $\RR^K$, while enforcing the orthogonality of such vectors to split them into separate groups. Therefore, the rows of $Y_\theta$ can be clustered with K-means to group together the nodes that are the most strongly connected by the representative graph.

Moreover, since the features of connected nodes are correlated (see our assumptions), the mapping actually learns to represent correlated features as close vectors of the latent space. This leads to a clustering-friendly representation of the whole feature space, because the learned mapping can be applied to any feature vector, and not only to those associated with the graph nodes. A generic feature vector $x\in\RR^M$ can be thus classified based on the distance of its embedding $f_\theta(x)$ to the cluster centers $\{c_1,\dots,c_K\}$ computed on the graph at training time, yielding the classifier defined as
\begin{equation}
(\forall x\in\RR^M)\quad p_{\rm class}(x) = \argmin{k\in\{1,\dots,K\}} \|f_\theta(x)-c_k\|.
\end{equation}

Fig.~\ref{fig:toy} presents an overview of the proposed framework.\footnote{The minimal requirement of our framework is to have at least a single-layer graph with an attribute for each node (in which case the first step is dropped), but additional layers and attributes are usually beneficial for the performance, as long as they are at least partially informative. In cases where the node features are not given, a one-hot indicator vector can be assigned to each node \cite{Schlichtkrull2018}. Conversely, if no graph is given, one or more layers can be built directly from the data, e.g., by computing several nearest-neighbor graphs on feature subsets. Although layers derived from the data might seem like redundant information, this is a common practice in image processing \cite{Chierchia2014, Wang2018, Tao2018} for example, where the nonlocal graph of similar patches is effectively used as a prior information to capture long-range correlations in the data.} 
More details about the two main steps of the proposed framework will be provided in the next sections.

	%--------------------------------------------------------------
	\section{Layer aggregation}
	\label{sec:spd}
	%--------------------------------------------------------------
	%
	%
	%
	\subsection{Problem formulation}

We now present the formulation for layer aggregation, which exploits the Riemannian geometry of SPD manifold to merge the Laplacian matrices $L^s$ of individual layers $\mathcal{G}^s$ into a single SPD matrix that describes the representative graph $G$. 

The notions of arithmetic and geometric means, typically used to average positive numbers, generalize naturally to a finite set of SPD matrices. This generalization is based on a variational characterization of the mean operation, which consists in finding the SPD matrix $L$ that minimizes its distance to a set of SPD matrices $\{L^s\}_{1\le s\le S}$, that is
\begin{equation}\label{eq:loss}
L = \Phi(L^1, \dots, L^S) = \argmin{L \in \mathcal{P}(N)} \sum_{s=1}^S \mathcal{D}(L, L^s),
\end{equation}
where $\mathcal{P}(N)$ denotes the manifold of $N\times N$ SPD matrices, and $\mathcal{D}\colon \mathcal{P}(N)\times\mathcal{P}(N) \to \RR$ is a suitable distance.

\subsection{Geometric mean}
When the dissimilarity between SPD matrices is computed via the Euclidean distance, the solution to Problem~\eqref{eq:loss} is the arithmetic mean of $\{L^s\}_{1\le s\le S}$. The latter is however suboptimal to merge information from different layers, and a better alternative for graph clustering is given by the matrix power mean \cite{MercadoGT018}, which can perfectly recover the clusters of complementary layers sampled from the stochastic block model when the power goes to $-\infty$.

A different family of matrix power means can be defined based on the Riemannian distance \cite{Lim2012}. In this context, the geometric mean arises as a special case of interest in machine learning \cite{Zadeh2016, Mercado2016}. The geometric mean of SPD matrices $\{L^s\}_{1\le s\le S}$ corresponds to the solution to Problem \eqref{eq:loss} with the Riemann distance\footnote{Note that the principal logarithm is not well defined on Laplacian matrices, as they are just positive semi-definite. To circumvent this issue, we add a small diagonal shift to ensure positive definiteness. In other terms, we implicitly assume that $L^s = \bar{L}^s + \epsilon I$, where $\epsilon$ is a small positive constant.} 
\begin{equation}\label{eq:riemann_dist}
\mathcal{D}(L,L^s) =
\big\|\Log\big(L^{-\frac{1}{2}}L^s L^{-\frac{1}{2}}\big)\big\|_F^2,
\end{equation}
where $\Log$ denotes the principal logarithm of a SPD matrix.

\subsection{Numerical computation}
We propose to aggregate the graph layers by computing the geometric mean of the respective Laplacian matrices. Formally, the geometric mean arises as the solution to Problem~\eqref{eq:loss} in the case when $\mathcal{D}$ is the Riemann distance given in~\eqref{eq:riemann_dist}. The problem however admits no close-form solution for $S>2$. We thus resort to numerically compute the geometric mean through the Fr\'echet-Karcher gradient flow \cite{karcher1977}, whose iterations are defined as follows
\begin{equation}
\label{eq:spd_manif}
(\forall t\in\mathbb{N})\quad L_{t+1} = L_{t}^{\frac{1}{2}} \textup{Exp} \Big(\beta \sum_{s=1}^S  \textup{Log} \big(L_{t}^{-\frac{1}{2}} L^s L_{t}^{-\frac{1}{2}}\big) \Big) L_{t}^{\frac{1}{2}}.
\end{equation}	
Here above, $L_0 = \sum_{s=1}^S L^s$ is the initialization, $\beta>0$ is the step size, and $\Exp$ denotes the exponential of a symmetric matrix. Riemannian gradient descent converges to the optimal solution with a rate of $O(1/k)$ for the geodesically convex problem considered here \cite{Zhang2016}. In practice, one iteration with $\beta=1$ suffices to find a good approximation of the solution. This operation has a time complexity of $O(N^3)$, due to the presence of matrix logarithms and exponentials.

\begin{figure}
	\subfloat[Individual layers composed by two complemetary blocks. The magnitude of edge weights is gray colored from white (small) to black (large). Node coloring is the result of spectral clustering with $K=2$ on each layer.\label{fig:spd:layers}]{\includegraphics[width=\linewidth]{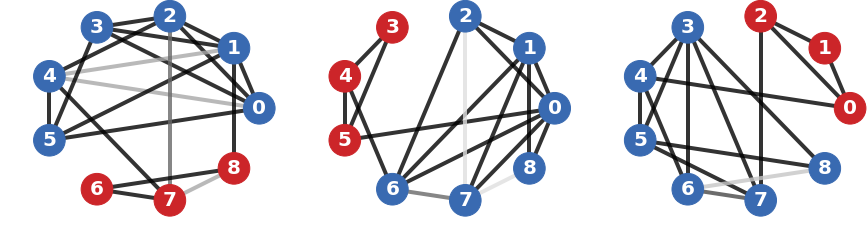}}
	\hfill
	\subfloat[Representative graphs obtained by aggregating the Laplacian matrices of individual layers with different techniques. \emph{Left:} Arithmetic mean. \emph{Middle:} Projection mean \cite{Dong_TSP_2014}. \emph{Right:} Geometric mean (proposed). Node coloring is the result of spectral clustering with $K=3$ on each aggregated graph. ]{\includegraphics[width=\linewidth]{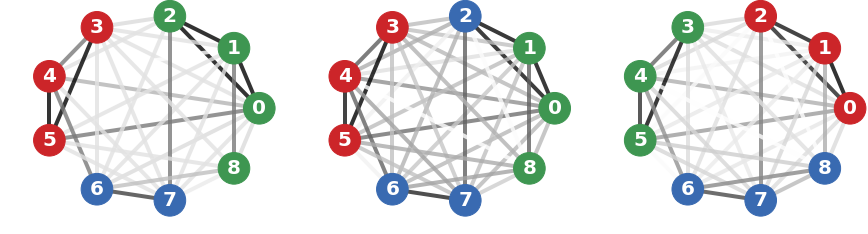}}
	\caption{Illustrative example of layer aggregation. The original graph is composed of three layers, which only provide partial information on the clustering structure. The geometric mean successfully merges the partial views into a representative graph with three distinct blocks (one from each layer).}
	\label{fig:spd}
\end{figure}

\subsection{Illustrative example}
Fig.~\ref{fig:spd} presents an example of layer aggregation, where the proposed geometric mean is compared to the arithmetic mean and the projection mean \cite{Dong_TSP_2014}. In this example, the original graph is composed of three layers, which only provide a partial information on the clustering structure. The geometric mean yields a graph that is representative of all the layers, as spectral clustering manages to recover from it the three blocks appearing in the respective layers. On the contrary, the arithmetic mean tends to underestimate the importance of edge $(7,8)$ despite that it appears in two layers, whereas the projection mean tends to overestimate the importance of edge $(0,8)$ despite that it appears in only one layer, leading to the incorrect assignment of nodes $2$ and $8$.

In the example of Fig.~\ref{fig:spd}, the merging of edges $(2,7)$, $(7,8)$, $(8,0)$, and $(6,8)$ is critical for correctly recovering the three clusters. Small changes of their weights result in different clustering for both the arithmetic and the projection mean. This is not the case for the geometric mean, which provides a consistent aggregation for a wide range of edge weights across different layers. Indeed, the Riemann distance is known to give equal importance to all eigenvalues, regardless of their magnitude. As a result, the geometric mean is more robust to small fluctuations of edge weights, and thus better suited to preserve the structural information of multilayer graphs.

	%--------------------------------------------------------------
	\section{Node feature embedding}
	\label{sec:embedding}
	%--------------------------------------------------------------
	%
	%
	%
	
\subsection{Problem formulation}

We now present the formulation for feature embedding, which relies on both the node attributes $X$ and the Laplacian matrix $L=[L_{ij}]$ of the representative graph $G$.

The goal is to estimate a mapping $f_\theta$ that represents the features of strongly connected nodes as close points in the latent space. That is, for large similarities $w_{ij} = -L_{ij}$, we want the distance between $f_\theta(x_i)$ and $f_\theta(x_j)$ to be small, which amounts to minimizing
\begin{equation}
\Tr(Y_\theta^\top L Y_\theta) = \sum_{i=1}^N \sum_{j=1}^N w_{ij} \|f_\theta(x_i) - f_\theta(x_j)\|^2,
\end{equation}
where $Y_\theta = f_\theta(X)$ is defined in \eqref{eq:Y_theta}. This objective is however not sufficient alone for learning an embedding that would result in effective clustering, as the sum of pairwise distances is trivially minimized by mapping all points to the same output vector. To avoid trivial solutions, we can borrow from spectral clustering the idea that the embedded points must be orthogonal to each other \cite{Shahaman2018,Pfau2019}, yielding
\begin{equation} \label{eq:embedding}
\minimize{\theta\in\RR^B}  \Tr(Y_\theta^\top L Y_\theta)
\quad{\rm s.t.} \quad 
Y_\theta^\top Y_\theta = {\bf I_{K\times K}}.
\end{equation}

The matrix $Y_\theta$ represents the graph nodes as vectors of the latent space $\RR^K$. Intuitively, nodes within the same cluster should be mapped to close vectors, while nodes from different clusters should be spaced out from each other, so that the latent space can be easily clustered. By optimizing the sum of pairwise distances over the orthogonality constraint, the rows of $Y_\theta$ tend to be split into $K$ clusters, which are formed by grouping together the more strongly connected vertexes in the graph. This is indeed similar to spectral clustering,\footnote{The connection to spectral clustering becomes apparent by setting $\theta\in\RR^{N\times K}$, $X={\bf I_{N\times N}}$, and $f_\theta(X) = X\theta$, so as to have $Y_\theta = \theta$.} which uses the spectrum of the Laplacian matrix to perform dimensionality reduction before clustering \cite{Malik2000,Ng2001,Luxburg2007}. An illustrative example of node feature embedding is presented in Fig.~\ref{fig:toy}, where the mapping $f_\theta$ is implemented by a neural net with four fully-connected layers and ReLU activations.

Note that many formulations have been proposed for representation learning on graphs \cite{hamilton2017}. There is however a clear distinction between classification scenarios, where metric learning methods are well established \cite{Chopra20005, Weinberger2009, Hoffer2015, Schroff2015, Bontonou2019}, and clustering scenarios. In the latter context, most approaches adopt a model based on two mapping functions: an \emph{encoder} that embeds each node into a low-dimensional vector, and a \emph{decoder} that recovers high-dimensional graph information (e.g., the node positions) from the learned embeddings. In particular, the decoder is needed for the definition of a self-supervised loss function that measures the discrepancy between the decoded similarity values and the true similarity values in the graph. 

The peculiarity of Problem \eqref{eq:embedding} is that the decoder is replaced by the orthogonality constraint. The advantage of this solution is that the mapping $f_\theta$ can directly learn the structural information provided by the graph, rather than indirectly using it through self-supervision. In addition, the mapping $f_\theta$ can be implemented by any parametric function from $\RR^M$ to $\RR^K$. This includes neural nets, whose only requirement is to end up with a layer of $K$ units. In practice, a small neural net with few fully-connected or graph-convolutional layers \cite{hamilton2017} is sufficient to effectively disentangle the data in a low-dimensional space. 
	\subsection{Proposed optimization algorithm} 
We propose to solve Problem~\eqref{eq:embedding} with an optimization algorithm based on gradient descent. The main difficulty of this optimization problem arises from the orthogonality constraint, since it is enforced on the mapping to be estimated, rather than the optimization parameters, ruling out standard techniques based on alternating optimization \cite{Lai2014}. While this problem was recently tackled in \cite{Shahaman2018, Pfau2019}, we propose an alternative algorithm based on implicitly constrained optimization, which extends our preliminary work \cite{elgheche2019_implicit}.

Our idea is that the orthogonality constraint in Problem \eqref{eq:embedding} can be enforced implicitly by using the upper triangular matrix $R_\theta^\top \in\RR^{K\times K}$ of the QR decomposition of $Y_\theta$, defined as
\begin{equation}
Y_\theta = Q_\theta R_\theta^\top,
\end{equation}
with $Q_\theta\in\RR^{N\times K}$ being a semi-orthogonal matrix. Indeed, when $Y_\theta$ is full rank, or equivalently $Y_\theta^\top Y_\theta$ is positive definite, its QR decomposition is unique, and $R_\theta$ is equal to the lower triangular factor of the Cholesky decomposition 
\begin{equation}
Y_\theta^\top Y_\theta = R_\theta R_\theta^\top. 
\end{equation}
Therefore, the semi-orthogonal factor $Q_\theta$ can be extracted from the matrix $Y_\theta$ by multiplying it with $R_\theta^{-\top}$, namely
\begin{equation}
Q_\theta = Y_\theta R_\theta^{-\top} \quad\Rightarrow\quad Q_\theta^\top Q_\theta =  {\bf I_{K\times K}}.
\end{equation}
This consideration allows us to rewrite Problem \eqref{eq:embedding} as
\begin{align} \label{eq:embedding_implicit}
\minimize{\theta\in\RR^B}{}  J(\theta) :=&\Tr(R_\theta^{-1} Y_\theta^\top L Y_\theta R_\theta^{-\top}) \\
\operatorname{with} \quad 
R_\theta :=& \chol(Y_\theta^\top Y_\theta).
\end{align}
In the above reformulation, the term $Y_\theta$ is no longer a semi-orthogonal matrix. The constraint is now enforced implicitly through the factor $R_\theta$ derived from the Cholesky decomposition of $Y_\theta^\top Y_\theta$, which ensures that the product $Y_\theta R_\theta^{-\top}$ is a semi-orthogonal matrix of $\mathbb{R}^{N\times K}$. This makes it possible to steer the embedding $Y_\theta$ away from trivial solutions.

\begin{algorithm}[t]
	\caption{Gradient descent for Problem \eqref{eq:embedding_implicit}}
	\label{algo:gradient_descent}
	\begin{algorithmic}[1]
		\Require{$L \in \RR^{N\times N}$}\Comment{Laplacian matrix}
		\Require{$X \in \RR^{N\times M}$} \Comment{Network data}
		\Require{$f_\theta\colon\RR^{M}\mapsto\RR^{K}$} \Comment{Neural net}
		\Require{$\theta_0\in\RR^B$} \Comment{Initialization}
		\Require{$\gamma > 0$} \Comment{Step size}
		
		\For{$t=0,1,\dots$}
		\State $Y_t = f_{\theta_t}(X)$
		\State $R_t = \chol(Y_t^\top Y_t)$
		\State $\overline{D}_t = R_t^{-T} R_t^{-1}$
		\State $\overline{Y}_t = 2\big(LY_t\overline{D}_t -Y_t \overline{D}_t Y_t^\top L Y_t \overline{D}_t\big)$
		\State $g_t = $ gradient of $J$ at $\theta_t$ based on $\overline{Y}_t$ \Comment{See \eqref{eq:gradient}}
		\State $\theta_{t+1} = \textrm{gradient-step}(\theta_t, g_t, \gamma)$ \Comment{See \cite{Reddi2018}}
		\EndFor
		\State \textbf{return} $f_{\theta_*}(X) R_*^{-\top}$
	\end{algorithmic}
\end{algorithm}

\subsection{Gradient descent}
All operations involved in Problem \eqref{eq:embedding_implicit} are differentiable, provided that the mapping $f_\theta$ is defined by a differentiable operator, such as a neural net. Specifically, the gradient of the cost function $J$ defined in \eqref{eq:embedding_implicit} can be decomposed as
\begin{equation}\label{eq:gradient}
\nabla J(\theta) = \left[\Tr\Big( \frac{\partial J(\theta)}{\partial Y_\theta}^\top\frac{\partial Y_{\theta}}{\partial \theta^{(b)}} \Big)\right]_{1\le b\le B}
\end{equation}
where the Jacobian w.r.t.\ $Y_{\theta}$ (derived in the appendix) reads
\begin{equation}\label{eq:jacobian}
\frac{\partial J(\theta)}{\partial Y_\theta} = 2\big({\bf I_{N\times N}} - Y_\theta R_\theta^{-T} R_\theta^{-1} Y_\theta^\top\big) L Y_\theta R_\theta^{-T} R_\theta^{-1}.
%LY_\theta R_\theta^{-T} R_\theta^{-1}
\end{equation}
Thanks to this result, a solution to Problem~\eqref{eq:embedding_implicit} can be found via gradient descent, whose iterations are summarized in Algorithm~\ref{algo:gradient_descent}. There are several advantages in solving Problem~\eqref{eq:embedding_implicit} with this approach. We avoid the complexity of alternating between a projection step and a gradient step \cite{Shahaman2018}, as the alternating approach may slow down the convergence to the optimal solution. Moreover, we can reduce the complexity of gradient updates via stochastic approximations \cite{Pfau2019, elgheche2019_implicit}. 

A question remains on the equivalence between the original problem \eqref{eq:embedding} and the proposed reformulation \eqref{eq:embedding_implicit}. By the Courant-Fischer theorem, the solution $Y_{\bar{\theta}}$ to Problem \eqref{eq:embedding} closely approximates the $K$ smallest eigenvectors of the matrix $L$, up to the representational capacity of the mapping $f_{\bar{\theta}}$. This is however not true for the solution $Y_{\bar{\theta}} R_{\bar{\theta}}^{-\top}$ derived from Problem~\eqref{eq:embedding_implicit}, which only spans the smallest $K$ eigenvectors of the matrix $L$ \cite{Edelman1999}. To see this, note that we can rewrite the cost function reformulated in \eqref{eq:embedding_implicit} as follows:
\begin{align}
\Tr(R_\theta^{-1} Y_\theta^\top L Y_\theta R_\theta^{-\top})
&= \Tr(R_\theta^{-\top} R_\theta^{-1} Y_\theta^\top L Y_\theta) \\
&= \Tr\big((R_\theta R_\theta^{-\top})^{-1} Y_\theta^\top L Y_\theta \big) \\
&= \Tr\big((Y_\theta^\top Y_\theta)^{-1} Y_\theta^\top L Y_\theta \big). \label{eq:quotient}
\end{align}
In the case $K=1$, this boils down to the Rayleigh quotient
\begin{align}
\mathcal{Q}(y) = \frac{y^\top L y}{y^\top y},
\end{align}
whose minimizer is  the smallest eigenvector of $L$. For $K>1$, the function \eqref{eq:quotient} is invariant to right-multiplications of $Y_\theta$, and thus the minimum is achieved by any basis that spans the $K$ smallest eigenvectors of $L$. This is not critical in our context, as a basis change in the embedding space does not affect clustering, allowing us to learn a valid mapping $Y_\theta$ without the need to explicitly compute the eigenvectors of $L$. Note also that we could directly minimize \eqref{eq:quotient} as in \cite{Pfau2019}. In this regard, the proposed cost function $J$ may be better suited for optimization, because it leads to a symmetric gradient, which improves stability during training.

	%\input{eigen.tex}

	%--------------------------------------------------------------
	\section{Experimental results}
	\label{sec:results}
	%--------------------------------------------------------------
	%
	%
	%

\subsection{Preliminaries}
We compare our approach with six clustering algorithms. The methods referred to as SC-ML \cite{Dong_TSP_2014}, MIMOSA \cite{Chen_Hero_2017}, and PLM \cite{MercadoGT018} work in two steps: they aggregate the Laplacian matrices of the multilayer graph, and then perform the spectral clustering of the resulting (single-layer) graph.\footnote{For a single-layer graph with $N$ nodes, a K-means clustering is computed on the rows of the matrix $U\in\RR^{N\times K}$ formed by the eigenvectors associated to the $K$ smallest eigenvalues of the graph Laplacian matrix, so as to group together the vertexes that are the most strongly connected by the graph.} The difference lies in the aggregation step, which is performed in Grassman manifold, via a convex combination, or using the power Laplacian mean, respectively. Moreover, the method called GMC \cite{Wang2019} jointly performs layer aggregation and spectral embedding, and the resulting embedding matrix is clustered with $K$-means. The method called SpectralNet \cite{Shahaman2018} builds a graph from the data points (features), estimates a nonlinear mapping by solving Problem~\eqref{eq:embedding}, and then embeds the features in a low-dimensional space, where they are clustered with K-means. We also report the performance of K-means clustering solely on the node features, without using the structural information carried by the multilayer graph.

All methods are used with their default hyperparameters. As for our method, referred to as OrthoNet, we implement the mapping as a neural net with four dense layers of size $400$-$200$-$100$-$K$ and PReLU activation \cite{He2015}, where $K$ is the number of desired clusters, and the optimization is carried out with AmsGrad method \cite{Reddi2018} using a learning rate $\gamma=10^{-3}$. We use three criteria to measure the clustering performance: Purity, Normalized Mutual Information (NMI), and adjusted Rand Index (RI). They measure the agreement of two partitions, ignoring permutations and with no requirement to have the same number of clusters. Values close to zero indicate two assignments that are largely independent, while values close to one indicate significant agreement. All the experiments are conducted in Python/Numpy/PyTorch on a 40-core Intel Xeon CPU at 2.5 GHz with 128GB of RAM.

\begin{figure}
	\centering
 	\includegraphics[width=\linewidth]{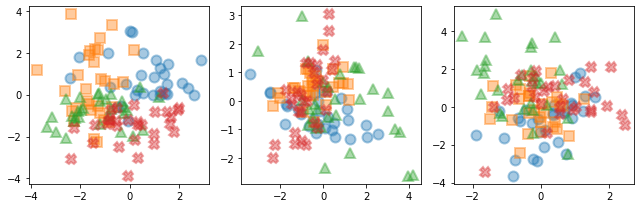}
 	\caption{Synthetic data generated with $N=100$ vectors, $K=4$ clusters, $S=3$ layers, and $d=2$ dimensions. Each panel shows the feature vectors of a layer, colored by the cluster they belong to. A k-NN graph is build on each set of vectors. The layer alignment is purely based on the clusters.}
 	\label{fig:synth_data}
 \end{figure}

\subsection{Datasets}\label{sec:datasets}

In our experiments, we consider several synthetic and real datasets. The synthetic dataset consists of $S$ point clouds of arbitrary size $N$, each generated from a $d$-dimensional Gaussian mixture model with $K$ components having different means and covariance matrices, as shown in Fig.~\ref{fig:synth_data}. We build a 20-nearest-neighbor ($k$-NN) graph on each point cloud, and we set the edge weights to the reciprocal of the Euclidean distance between pairs of neighbors. This give us $S$ layers. Then, we concatenate the data points across the clouds, so as to form a feature vector of size $M=dS$ for each node in the multilayer graph, where the alignment of nodes across layers is known by construction. The goal is to recover the $K$ components from which the data points are generated.

We then consider several real datasets. The IMDB database allows access to the movie's actors, directors, writers and production company, the movie's awards (wins and nominations), its box office, as well as the directors/actors/writers box office. Without having access to budget figures, the goal is to cluster the movies into $K=5$ budget ranges: low cost (below $0.1$ USD millions), low-medium cost (below 10 USD millions), medium cost (below 40 USD millions), medium-high cost (below 100 USD millions), high cost (above 100 USD millions). To build a multilayer graph on IMDB data, we connect the movies sharing one or more actors, directors, or writers, leading to $S=3$ graph layers. Moreover, each movie is associated to $M=3$ attributes: box office, awards, and director box office.

Moreover, Yelp is a popular website for reviewing and rating local businesses. In our experiments, we only extract star ratings, text reviews, and review evaluations (users can mark reviews as ``cool'', ``useful'', and ``funny''), ignoring the other information in the dataset. The goal is to cluster the businesses into $K=3$ rating levels: low (1 or 2 stars), medium (3 stars), high (4 or 5 stars). To build a multilayer graph on Yelp data, we proceed as follows. The text reviews are preprocessed using sentiment analysis. This yields a polarity score within the range $[-1, 1]$ on which we build a $20$-NN graph. We also build a $20$-NN graph on the review evaluations, leading to $S=2$ graph layers. Moreover, each business is associated to $M=2$ attributes: the sentiment analysis score, and the review evaluation score.

Another dataset, the ``100 leaves'', contains $N=1600$ samples of $M=192$ features for $K=100$ plant species. We build a $5$-NN graph on $S=3$ different feature subsets: shape descriptor, fine scale margin, and texture histogram. The goal is to cluster the observations according to their plant species.

Finally, the ``Mfeat'' handwritten digit dataset contains $N=2000$ samples of $M=650$ features for $K=10$ digits  (0-9). We build a $5$-NN graph on $S=6$ different feature groups. The goal is to cluster the observations according to their digits.

\begin{table*}
	\centering
	\caption{One-layer graph clustering versus multilayer graph clustering \label{t:table1}}
	
	\begin{tabular}{ c c | c c c | c c c }
		
		\toprule
		
		& & \multicolumn{3}{c}{\textsc{Spectral clustering}} & \multicolumn{3}{c}{\textsc{OrthoNet}}\\
		& & \multicolumn{3}{c}{(graph only)} & \multicolumn{3}{c}{(graph + features)}\\
		Dataset & Layer aggregation &  Purity & NMI & RI & Purity & NMI & RI \\
		\midrule
		\underline{\textsc{Synthetic}} &&&&&
		\\
		%\cmidrule(lr){0-0}
		Nodes: 2000 -- Clusters: 5 & None & 0.9375 & 0.8626 & 0.8568 & 0.9395 & 0.8647 & 0.8610
		\\
		\textbf{Layers: 1} -- Features: 8 &&&&&
		\\
		\midrule
		
		\underline{\textsc{Synthetic}} & Average & 0.9475 & 0.8437 & 0.8731 & 0.9535 & 0.8614 & 0.8873
		\\
		%\cmidrule(lr){0-0}
		Nodes: 2000 -- Clusters: 5 & SC-ML & 0.9555 & 0.8634 & 0.8919 & 0.9580 & 0.8718 & 0.8977
		\\
		\textbf{Layers: 4} -- Features: 8 & SPD & 0.9705 & 0.9075 & 0.9275 & 0.9720 & 0.9113 & 0.9312
		\\
		\midrule
		
		\underline{\textsc{Mfeat}} &&&&&
		\\
		%\cmidrule(lr){0-0}
		Nodes: 2000 -- Clusters: 6 & None & 0.7900 &  0.7489 & 0.6558 & 0.8000 &  0.7631 & 0.6752
		\\
		\textbf{Layers: 1} -- Features: 650 &&&&&
		\\
		\midrule
		
		\underline{\textsc{Mfeat} } & Average & 0.8395 & 0.8410 &  0.7622  &  0.8515& 0.8371 & 0.7761 
		\\
		Nodes: 2000 --  Clusters: 6 & SC-ML & 0.8775 & 0.8780 & 0.8339 & 0.8920  &  0.8860 & 0.8438
		\\
		\textbf{Layers: 6} -- Features: 650 & SPD & 0.9130 & 0.8953 & 0.8575  & 0.9555 & 0.9128 & 0.9057
		\\	  
		\bottomrule
	\end{tabular}
\end{table*}

\begin{table*}
	\centering
	\caption{Performance of multilayer graph clustering algorithms. \label{t:table2}}
	
	\begin{tabular}{ l   c @{\quad}   c @{\quad} c @{\quad} c @{\quad}  c @{\quad} c @{\quad} | c }
		
		\toprule
		
		Data 
		& K-means &SC-ML & GMC & PLM & MIMOSA 
		& {\footnotesize SpectralNet}  & OrthoNet \\
		\midrule
\\

\multirow{1}{*}{\textsc{Synthetic}} & \multicolumn{7}{l}{ (Nodes: 10000, Features: 8, Layers: 4, Clusters: 5) }\\	 
\\ 

Purity         & 0.9901 & 0.9858   & 0.3996 & 0.9886            & 0.9868   & \textbf{0.9940}  &  \textbf{0.9948}\\      
NMI            &  0.9664 & 0.9545   & 0.7299 & 0.9621            & 0.9566   &  \textbf{0.9720} &   \textbf{0.9756} \\   
RI               &  0.9756 & 0.9652   & 0.4763 & \textbf{0.9910}   & 0.9896   & {0.9840}         &   {0.9827}  \\
Time          & 0.10 s & 153.84 s & 1345 s & 408.62 s          & 1418 s   & 464.31 s         &   785.84 s  \\ %6.56 s \\ % 
\\
\multirow{1}{*}{\textsc{YELP} } & \multicolumn{7}{l}{ (Nodes: 1600, Features: 2, Layers: 2, Clusters: 3) }\\
\\
Purity         & 0.8656 & 0.7748  & 0.7007 &    0.7616	& 0.7168   &  0.9460 & \textbf{0.9570}  \\         
NMI           & 0.6265 & 0.3329  & 0.3925 &   	0.6754	& 0.1395   &  0.6810  & \textbf{0.7910}  \\   
RI             & 0.5912 &  0.5192 & 0.6010 &    0.8433  &  0.6603    &  0.7322  & \textbf{0.9044}  \\   
Time          & 0.016 s &  1.54 s & 7.3 s  &    5.10 s  & 115.67 s   &  82.04 s & 6.42 s \\ %1.75 s \\ 
\\   
\multirow{1}{*}{\textsc{IMDB} } & \multicolumn{7}{l}{ (Nodes: 550, Features: 3, Layers: 3, Clusters: 5) }\\
\\ 
Purity        & 0.8280  & 0.8817  & 0.7007 & 0.8495          &  0.5358    &   0.7310  &  \textbf{0.8925} \\      
NMI          &  0.2331 & 0.3614  & 0.1563 & 0.2260          &  0.0953    &   0.2200  &  \textbf{0.5056} \\      
RI             & 0.0219 & 0.2680  & 0.5245 & \textbf{0.7437} &  0.4347    &   0.1810  &  0.6909 \\      
Time         & 0.04 s & 0.31 s  & 2.21 s & 20.89 s         &  50.6 s   &  30.35 s &  19.68 s\\ %1.97 s \\         
\\     
\multirow{1}{*}{\textsc{100 Leaves} } & \multicolumn{7}{l}{ (Nodes: 1600, Features: 192, Layers 3, Clusters: 100) }\\
\\
Purity         & 0.6987 &  0.9487 & 0.8237 & 0.8229 &   --    & 0.7840 & \textbf{0.9712} \\      
NMI            & 0.8452 &  0.9717 & 0.9292 & 0.9079 &   --    & 0.8370 & \textbf{0.9812}  \\     
RI               & 0.5578 &  0.9129 & 0.4974 & \textbf{0.9819} &   --    & 0.3530 & 0.9478  \\    
Time          & 1.12 s&  1.86 s & 8.09 s & 4.06 s &   --    & 770 s  & 4.17 s  \\    
\\     
   
\multirow{1}{*}{\textsc{Mfeat} } & \multicolumn{7}{l}{ (Nodes: 2000, Features: 650, Layers: 6, Clusters: 10) }\\
\\
Purity         & 0.5470 & 0.8775  & 0.8820  & 0.8780  & 0.2215  & 0.7685 & \textbf{0.9555} \\      
NMI          &  0.5744 & 0.8780  & 0.9041  & 0.8807  & 0.3549  & 0.7480  & \textbf{0.9128}  \\     
RI             &  0.4293 & 0.8339  & 0.8496  & 0.9692  & \textbf{0.9960}  & 0.6343  & 0.9057  \\    
Time         & 0.71 s  & 3.45 s  & 24.05 s & 11.82 s & 13.04 s &   95.12 s  &   10.85 s \\    

		\bottomrule
	\end{tabular}
\end{table*}

\begin{figure*}
	\centering
	\subfloat[Synthetic]{\includegraphics[width=0.3\linewidth]{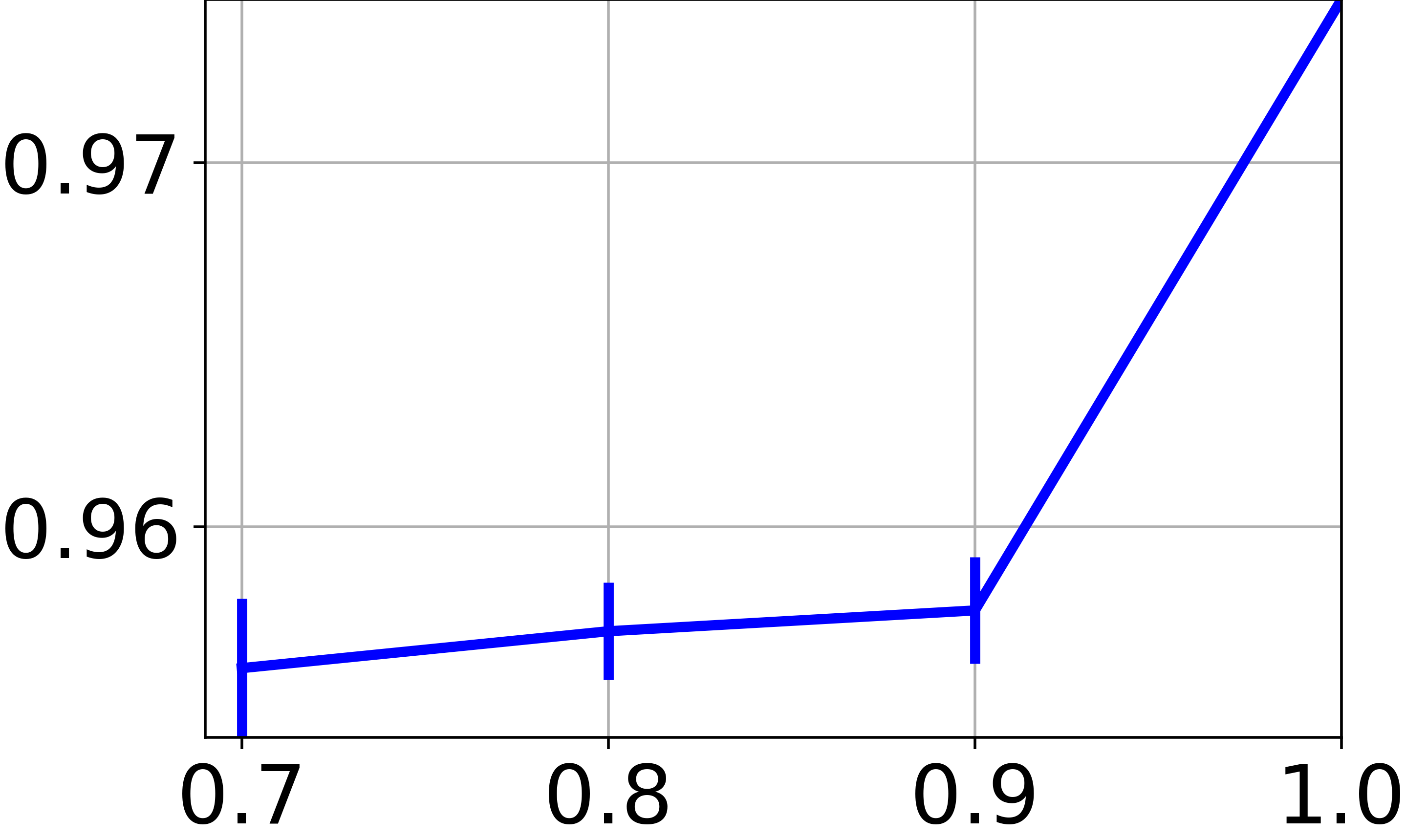}}
	\hfill
	\subfloat[YELP]{\includegraphics[width=0.3\linewidth]{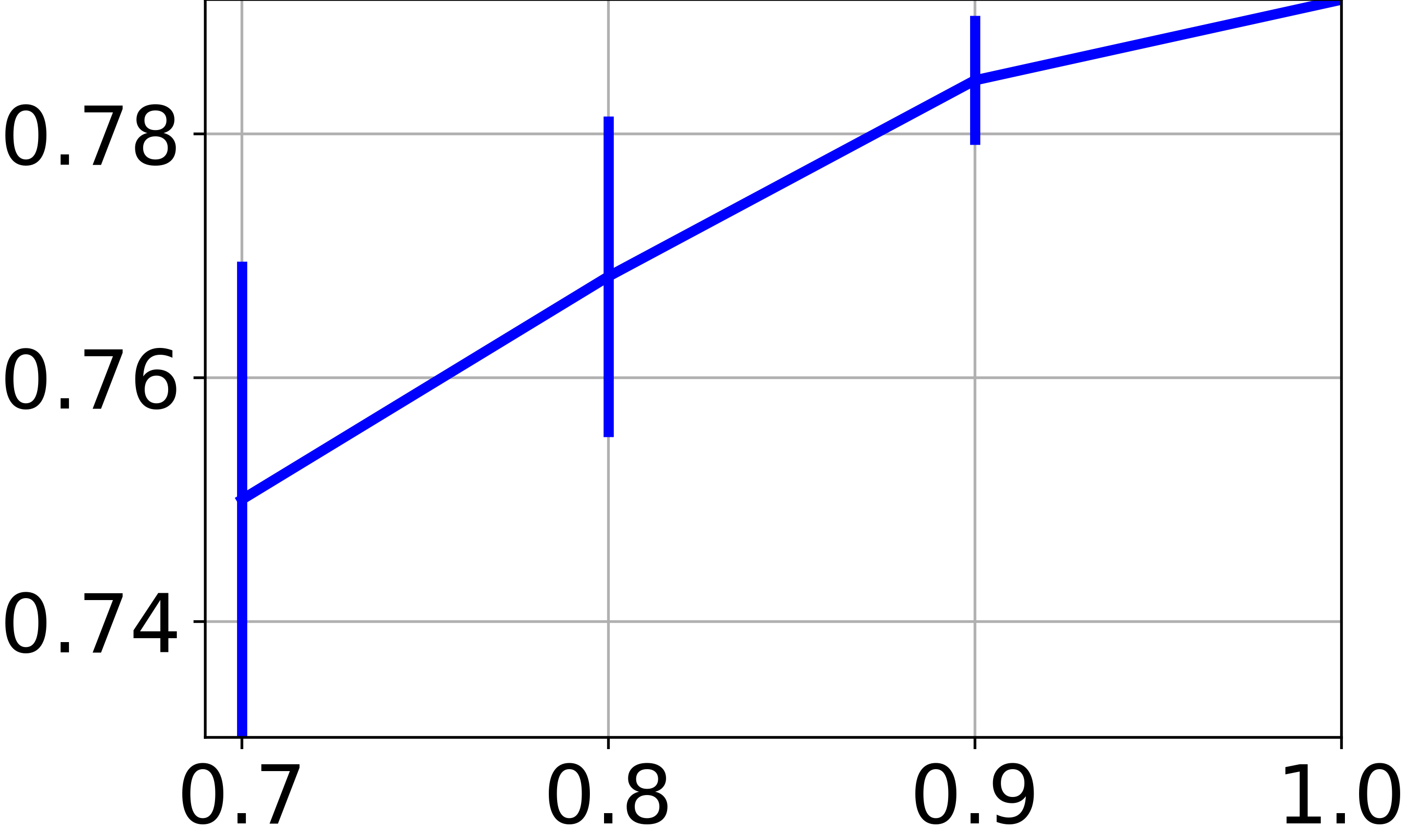}}
	\hfill
	\subfloat[IMDB]{\includegraphics[width=0.3\linewidth]{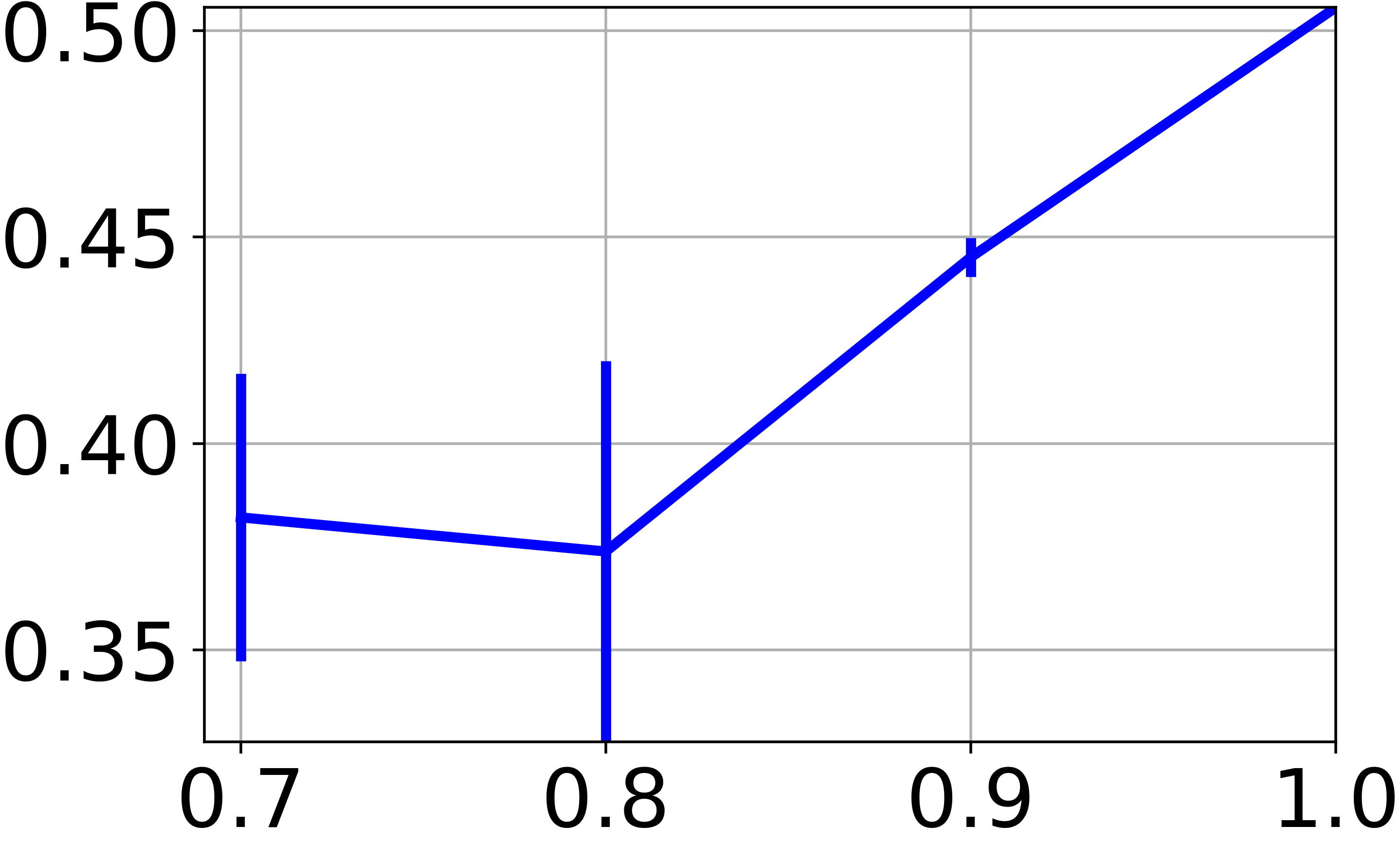}}
	
	\hfill
	\subfloat[100 leaves]{\includegraphics[width=0.3\linewidth]{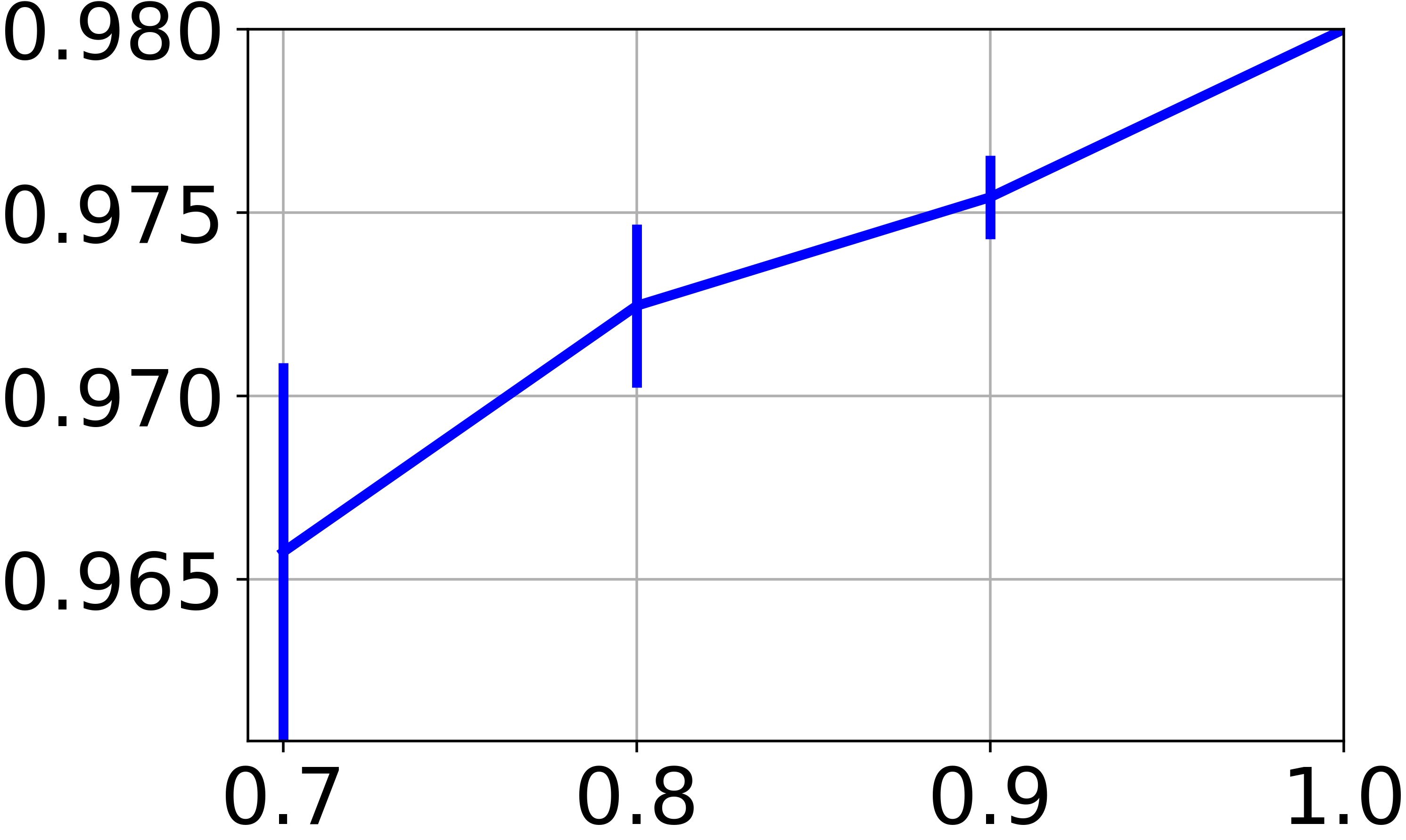}}
	\hfill
	\subfloat[Mfeat]{\includegraphics[width=0.3\linewidth]{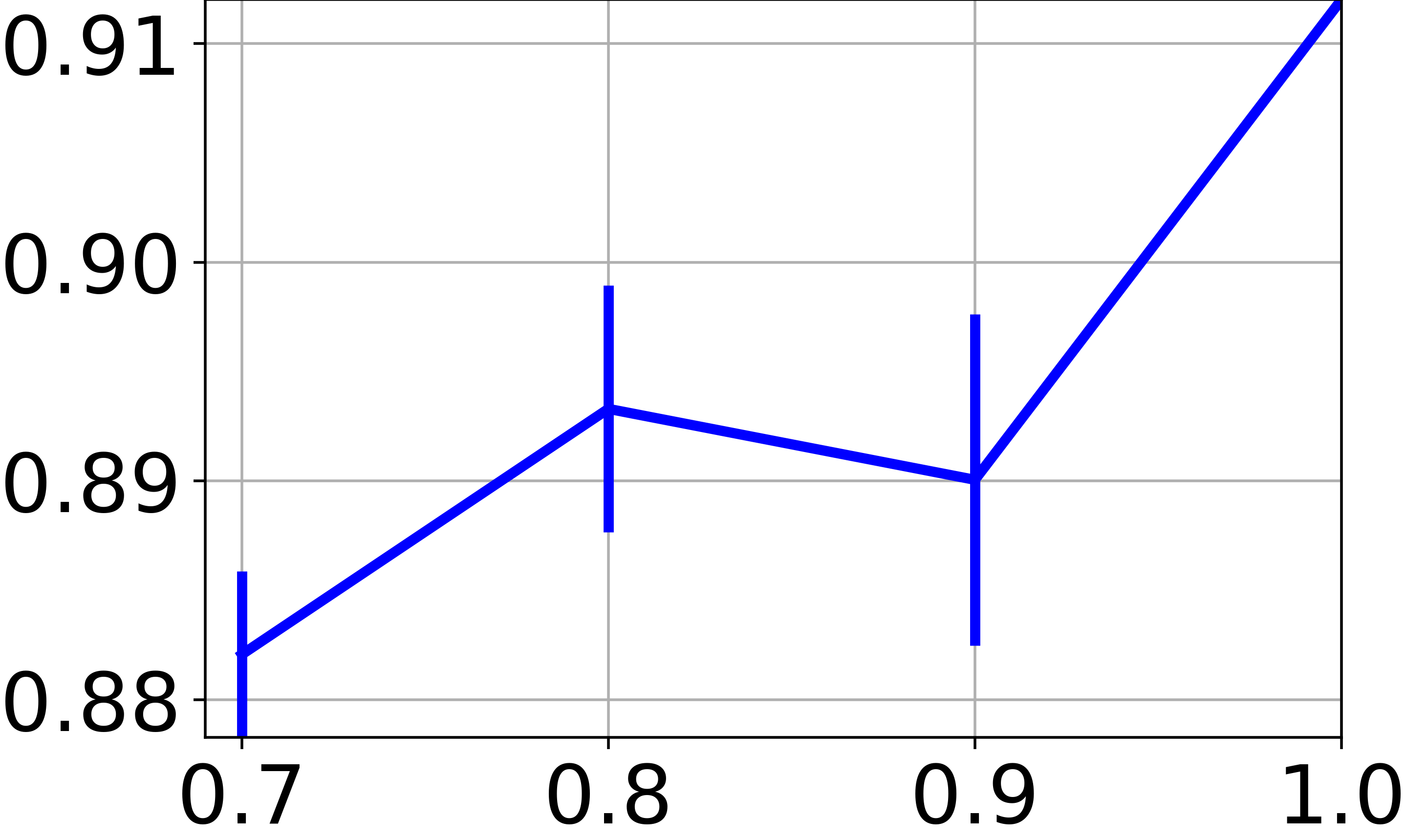}}
	\hfill\mbox{}
	\caption{OrthoNet performance (y-axis) in terms of NMI evaluated on all the data, after the training is performed on a fraction (x-axis) of data and graph.}
	\label{fig:general} 
\end{figure*}

\subsection{Graph clustering: single layer versus multiple layers}
We start our analysis by assessing the importance of building a multilayer graph on several subsets of features, as opposed to constructing a one-layer graph on the whole features. To this end, we compare two alternative strategies to build the graph on a given dataset.
\begin{itemize}
	\item \emph{Single-layer graph}. For each sample, we concatenate all the available features into one vector. Then, we use those vectors to build a single k-NN graph.
	\item \emph{Multilayer graph}. For each sample, we split the available features into different subsets (see Section \ref{sec:datasets}). Then, we build a separate k-NN graph on each subset, and we stack them into a multilayer graph.
\end{itemize}

Table~\ref{t:table1} reports the clustering performance on two datasets (synthetic and Mfeat) using the above graph building strategies. We directly perform spectral clustering and OrthoNet clustering on single-layer graphs, leading to the two sets of indicators reported in the first and third line of Table~\ref{t:table1}. On multilayer graphs, we first aggregate the layers using three different methods: the arithmetic mean (average), the projection mean in Grassman manifold (SC-ML), and the geometric mean in SPD manifold (proposed). Then, we perform spectral clustering or OrthoNet clustering on each aggregated graph, leading to the six sets of indicators reported in the second and fourth line of Table~\ref{t:table1}. In both scenarios, spectral clustering  relies only on the graph, whereas OrthoNet relies on both the graph and the features.

Among the results obtained with spectral clustering (left side of the table) and OrthoNet clustering (right side of the table), the proposed approach (SPD+OrthoNet) achieves the best performance. This empirically confirms that 
\begin{itemize}
	\item multilayer graphs carry a richer information than single-layer graphs for the purpose of node clustering, 
	\item the geometric mean in SPD manifold is an appropriate and effective choice for layer aggregation, 
	\item the presence of features on graph nodes can improve the clustering of multilayer graphs. 
\end{itemize}

\subsection{Multilayer graph clustering: performance assessment}
Table \ref{t:table2} reports a broader comparison with the state-of-the-art methods mentioned in the previous subsection.\footnote{We were unable to run MIMOSA on ``100 leaves'' dataset, due to the high number (100) of clusters.} On the synthetic dataset, OrthoNet and SpectralNet are the best performers, whereas the aggregation-based techniques are practically equivalent. This may be related to the fact that signals are more relevant for this kind of data, than the graphs alone (which are built from the signals). On the other datasets, OrthoNet is by far the best performer, especially in terms of the NMI score. This result is probably due to the richer information carried by graph layers, which cannot be translated into features. As SpectralNet builds a similarity distance on the feature vectors (using nearest neighbors or Siamese network), it cannot rely on the benefit brought by the multilayer graph. Conversely, the proposed approach can take advantage of the information carried by both the multilayer graph and the feature vectors, leading to a better clustering performance. Such improvement is however achieved with an increase of the execution time, due to the high computational cost for computing the geometric mean of SPD matrices.

\subsection{Generalization to new data}\label{sec:generalization}
In our approach, we train a neural net to find a clustering-friendly representation of the feature space. While so far we focused on the clustering of graph node features, we now wish to assess the capacity to classify new feature vectors never seen before (i.e., not associated to any graph node). The difficulty here is that we deal with a completely unsupervised scenario, so the experimentation protocol based on splitting the data in train and test sets is not really meaningful. To allow for a fair comparison with the techniques reported in Table \ref{t:table2}, we train $f_{\theta}$ on a subset of the available graph nodes, and then we evaluate the clustering performance of $f_{\theta}$ on all the available feature vectors. In particular, Fig.~\ref{fig:general} reports the NMI score obtained by training OrthoNet on $70\%$, $80\%$, $90\%$, and $100\%$ of the graph nodes. For all the fractions less than $100\%$, we repeat the training 10 times on random subsets of the graph, and we report both mean and standard deviation of the scores. The results show a moderate drop of performance when training is performed of smaller graphs. This suggests that the learned neural net has the ability to generalize to new data, provided that the graph carries sufficient information.

	%--------------------------------------------------------------
	\section{Conclusion}
	\label{sec:conclusion}
	%--------------------------------------------------------------
	%
	%
	%
	We have proposed a framework for multilayer network data clustering based on a two-step approach. We first compute the geometric mean of Laplacian matrices in the SPD manifold, and then we use the resulting graph to train a neural net on the node features in a unsupervised manner, using a formulation inspired from spectral clustering. The latter step is tackled with a new optimization algorithm that deals with the orthogonality constraint of the neural net outputs in an implicit way, so as to span the leading eingenvectors of the aggregated Laplacian matrix without the need to explicitly compute them. Experimental results show a better clustering performance of this approach on diverse datasets compared to state-of-the-art multilayer network clustering, as well as the ability of the trained neural net to generalize to new data. Interesting perspectives for future work include a better modeling of node features through a general approach to simultaneously aggregate the multilayer information with the network data.

	%--------------------------------------------------------------
	\appendix[Derivation of the gradient]
	\label{sec:gradient}
	%--------------------------------------------------------------
	%
	%
	%
	
\begin{table*}
\centering
\caption{Step-by-step reverse-mode differentiation}
\label{tab:reverse_mode}
\newcommand{\dif}[1]{{\rm d}#1}
\begin{tabular}{llll}
\toprule
Operation & Differential of a variable & Differential of $J$ & Jacobian of $J$ w.r.t. a variable \\
\midrule
$J = \Tr(C)$ & - & $\dif{J} = \Tr(\overline{C}^\top dC )$ & $\overline{C} = I$ \\
$C = A D A^\top$ & $\dif{C}=\dif{A} \, D A^\top + A \, \dif{D} \, A^\top + A  D \, \dif{A}^\top$ & $\dif{J} = {\rm Tr}\big(\overline{A}^\top \dif{A}  + \overline{D}^\top \dif{D}\big)$ & $\overline{A} = 2 \, A D$\\ 
&&& $\overline{D} = A^\top A$\\
$D = Y^\top L Y$ & $\dif{D}=\dif{Y}^\top L Y + Y^\top L \, \dif{Y}$ & $\dif{J} = {\rm Tr}\big(\overline{A}^\top \dif{A}  + \widetilde{Y}^\top \dif{Y}\big)$ & $\widetilde{Y} =  2 \, L Y \overline{D}$ \\
$A = R^{-1}$ & $\dif{A}= -A \, \dif{R} \, A$ & $\dif{J} = {\rm Tr}\big(\overline{R}^\top \, \dif{R} + \widetilde{Y}^\top \dif{Y} \big)$  & $\overline{R} = -R^{-\top} \overline{A} R^{-\top}$ \\
$R = {\rm cholesky}(P)$ & $\dif{R}=R \, \Phi\big(R^{-1} \dif{P} \, R^{-\top}\big)$ & $\dif{J} = {\rm Tr}\big(\overline{P}^\top \dif{P} + \widetilde{Y}^\top \dif{Y} \big)$ & $\overline{P} = \frac{1}{2}(\mathcal{S} + \mathcal{S}^\top)$ \\
&&& $\mathcal{S} = R^{-\top}\Phi(R^\top\overline{R})R^{-1}$ \\
&&& $\Phi(\cdot) = \cdot - {\rm triu}(\cdot) + \frac{1}{2} {\rm diag}(\cdot)$\\
$P = Y^\top Y$ & $\dif{P}=\dif{Y}^\top Y + Y^\top \dif{Y}$ & $\dif{J} = {\rm Tr}\big(\overline{Y}^\top \dif{Y}\big)$ & $\overline{Y} = 2 \, Y \overline{P} + \widetilde{Y}$\\
\bottomrule
\end{tabular}
\end{table*}

To derive the gradient of the cost function $J$ defined in \eqref{eq:embedding_implicit}, we first apply the chain rule and obtain the expression in \eqref{eq:gradient}. Then, the Jacobian of $J$ w.r.t.\ $Y_\theta$ can be derived by reverse-mode algorithmic differentiation \cite{Giles2008, Murray2016}. The step-by-step computation is reported in Table~\ref{tab:reverse_mode}, where the standard algorithmic differentiation terminology is used: if the matrix $A$ is an intermediate variable within the cost function $J$, then $\overline{A}$ denotes the derivative of $J$ w.r.t.\ each element of $A$. From Table~\ref{tab:reverse_mode}, we deduce that the derivative w.r.t. $Y_\theta$ reads
\begin{equation}\label{eq:gradient_proof_step1}
\overline{Y}_\theta = 2 \, Y_\theta \overline{P} + 2 L Y_\theta R_\theta^{-T}R_\theta^{-1}
\end{equation}
where
\begin{align}
\overline{P} &= -\frac{1}{2} R_\theta^{-\top} \big( \Phi(2R_\theta^{-1} D R_\theta^{-T}) + \Phi(2R_\theta^{-1} D R_\theta^{-T})^\top \big)R_\theta^{-1} \nonumber \\
&= -R_\theta^{-\top} R_\theta^{-1} D R_\theta^{-T} R_\theta^{-1} \nonumber \\
&= -R_\theta^{-\top} R_\theta^{-1}  Y_\theta^\top L Y_\theta R_\theta^{-T} R_\theta^{-1}.
\label{eq:gradient_proof_step2}
\end{align}
By putting together \eqref{eq:gradient_proof_step1} and \eqref{eq:gradient_proof_step2}, we arrive at the final expression of the Jacobian given in \eqref{eq:jacobian}.

	%--------------------------------------------------------------
	\bibliographystyle{IEEEtran}
	\bibliography{IEEEabrv,biblio}
	%--------------------------------------------------------------

\end{document}